\RequirePackage{amsmath}
\documentclass[runningheads]{llncs}
\usepackage{amsmath,amssymb}
\usepackage{graphicx}
\usepackage{color}
\usepackage[width=122mm,left=12mm,paperwidth=146mm,height=193mm,top=12mm,paperheight=217mm]{geometry}

\usepackage[blocks]{authblk}
\usepackage{array}
\usepackage{times}
\usepackage{epsfig}
\usepackage{xcolor}
\usepackage{acronym}
\usepackage{rotating}  
\usepackage{xspace}
\usepackage{changes}  

\usepackage[percent]{overpic}

\begin{document}


\def\figref#1{Fig.~{\ref{#1}}}

\newcommand{\etal}{\textit{et al}.}
\newcommand{\ie}{\textit{i}.\textit{e}.}
\newcommand{\eg}{\textit{e}.\textit{g}.}

\newcommand{\mypara}[1]{\paragraph{\textbf{#1.}}}

\def\facade{fa\c{c}ade\xspace}
\def\facades{fa\c{c}ades\xspace}
\def\Facade{Fa\c{c}ade\xspace}
\def\segnet{\textsc{Segnet}\xspace}
\def\Segnet{\segnet}

\def\segnetfacade{\textsc{FacSegnet}\xspace}
\def\segnetmultifacade{\textsc{MultiFacSegnet}\xspace}
\def\separable{\textsc{Separable}\xspace}  
\def\segnetseparable{\separable{}}
\def\segnetcompatability{\textsc{Compatibility}\xspace}

\def\eTRIMS{eTRIMS\xspace}
\def\numeTRIMSimages{60\xspace}

\def\softmax{softmax\xspace}

\def\numlabels{11\xspace}
\def\numholdouts{293\xspace}
\def\numcmpholdouts{175\xspace}
\def\numourholdouts{118\xspace}
\def\numcmpimages{606\xspace}
\def\googlestreetview{\ac{GSV}\xspace}

\def\photosphere{photosphere\xspace}
\def\openstreetmap{OpenStreetMap\xspace}

\def\metersperpixel{0.025}  
\def\nummetersaway{20} 
\def\nummeterstall{40} 
\def\nummeterslong{40} 
\def\mergetolerance{2} 

\def\nummetersexteded{2}
\def\meters{m\xspace}

\def\warpamount{20\xspace}  

\def\camvid{Camvid\xspace}
\def\Camvid{Camvid\xspace}

\def\Target#1{\text{\texttt{{#1}}}}
\def\windowlabel{\Target{window}\xspace}
\def\balconylabel{\Target{balcony}\xspace}
\def\treelabel{\Target{tree}\xspace}
\def\skylabel{\Target{sky}\xspace}
\def\shoplabel{\Target{shop}\xspace}

\acrodef{NEG}[\Target{NEG}]{negative}
\acrodef{POS}[\Target{POS}]{positive}
\acrodef{EDG}[\Target{EDG}]{edge}
\acrodef{UNK}[\Target{UNK}]{unknown or unlabeled}

\def\NEG{{\ac{NEG}}}
\def\POS{{\ac{POS}}}
\def\UNK{{\ac{UNK}}}
\def\EDG{{\ac{EDG}}}

\def\pxprecision{\ensuremath{P}}
\def\pxrecall{\ensuremath{R}} 
\def\pxfscore{\ensuremath{F_1}}
\def\pxacc{\ensuremath{\mathit{Acc}}}
\def\pxTP{\ensuremath{\mathit{TP}}\xspace}
\def\pxFP{\ensuremath{\mathit{FP}}\xspace}
\def\pxFN{\ensuremath{\mathit{FN}}\xspace}
\def\pxTN{\ensuremath{\mathit{TN}}\xspace}
\def\pxExpected{\ensuremath{\mathit{Expected}}\xspace}
\def\pxPredicted{\ensuremath{\mathit{Predicted}}\xspace}

\def\obprecision{\ensuremath{P_\mathit{ob}}}
\def\obrecall{\ensuremath{{R}_\mathit{ob}}}
\def\obfscore{\ensuremath{{F_1}_\mathit{ob}}}
\def\obExpected{\ensuremath{\mathit{Expected}_\mathit{ob}}\xspace}
\def\obPredicted{\ensuremath{\mathit{Predicted}_\mathit{ob}}\xspace}
\def\obTP{\ensuremath{\mathit{TP}_\mathit{ob}}\xspace}
\def\obFP{\ensuremath{\mathit{FP}_\mathit{ob}}\xspace}
\def\obFN{\ensuremath{\mathit{FN}_\mathit{ob}}\xspace}

\def\intersect{\cap}

\acrodef{CRF}{Conditional Random Field}
\acrodef{IoU}{Intersection over Union}
\acrodef{AP}{Average Precision across classes}
\acrodef{mAP}{mean Average Precision} 
\acrodef{GSV}{Google Streetview}
\acrodef{CNN}{Convolutional Neural Network}
\acrodef{GPS}{Geographic Positioning System}
\acrodef{GIS}{Geographic Information System}
\acrodef{SGD}{Stochastic Gradient Descent}

\def\degrees{$^\circ$}


\pagestyle{headings}

\mainmatter
\def\ECCV18SubNumber{2618}  

\title{Facade Segmentation in the Wild}

\titlerunning{Facade Segmentation in the Wild}
\authorrunning{Femiani \textit{et al.}}


\author{
    John Femiani\textsuperscript{1} \hspace{3mm}
    Wamiq Reyaz Para\textsuperscript{2} \hspace{3mm}
    Niloy Mitra\textsuperscript{3} \hspace{3mm}
    Peter Wonka\textsuperscript{2} \hspace{3mm}\\
     Miami University\textsuperscript{1} \hspace{5mm}
     KAUST\textsuperscript{2} \hspace{5mm}
     UCL\textsuperscript{3} 
}
\institute{}

\maketitle


\begin{abstract} 
Urban \facade segmentation from automatically acquired imagery, in contrast to traditional image segmentation, poses several unique challenges. 360$^\circ$ photospheres captured from vehicles are an effective way to capture a large number of images, but this data presents difficult-to-model warping and stitching artifacts. In addition, each pixel can belong to multiple \facade elements, and different facade elements (e.g., \Target{window}, \Target{balcony}, \Target{sill}, etc.) are correlated and vary wildly in their characteristics. 
In this paper, we propose three network architectures of varying complexity to 
achieve  multilabel semantic segmentation of \facade images while exploiting their unique characteristics.  
Specifically, we propose a \segnetmultifacade architecture to assign multiple labels to each pixel, a \separable  architecture as a low-rank formulation that encourages extraction of rectangular elements, and a \segnetcompatability network that simultaneously seeks segmentation across facade element types allowing the network to `see' intermediate output probabilities of the various \facade element classes. 
%
Our results on benchmark datasets show significant improvements over existing \facade segmentation approaches for the typical \facade elements. For example, on one commonly used dataset the accuracy scores for \Target{window} (the most important architectural element) increases from 0.91 to 0.97 percent compared to the best competing method, and  comparable improvements on other element types.

\keywords{\facade\ segmentation, learning architecture, semantic segmentation}

\end{abstract}

\acresetall

\section{Introduction}
\label{sec:introduction}

We propose a deep learning based solution for the per-pixel semantic segmentation and classification of \facade\ images.
Although many methods, both hand-crafted and deep learning based, exist for semantic image  segmentation, \facade images have certain special characteristics that prevent  state-of-the-art semantic image  segmentation methods from being directly used in our setting.
Note that unlike typical semantic segmentation, \facade  per-pixel  segmentation is special as a pixel can be {\em simultaneously} assigned to multiple labels (e.g., \Target{window} and \Target{balcony}).
Further, many \facade\ labels are very thin (e.g., \Target{sill} or \Target{pillar}) and some features such as the partitions between adjacent buildings require special handling for reliable detection. 

\begin{figure*}[ht]
    \centering
    \def\myheight{1.9in}
    \includegraphics[width=\textwidth]{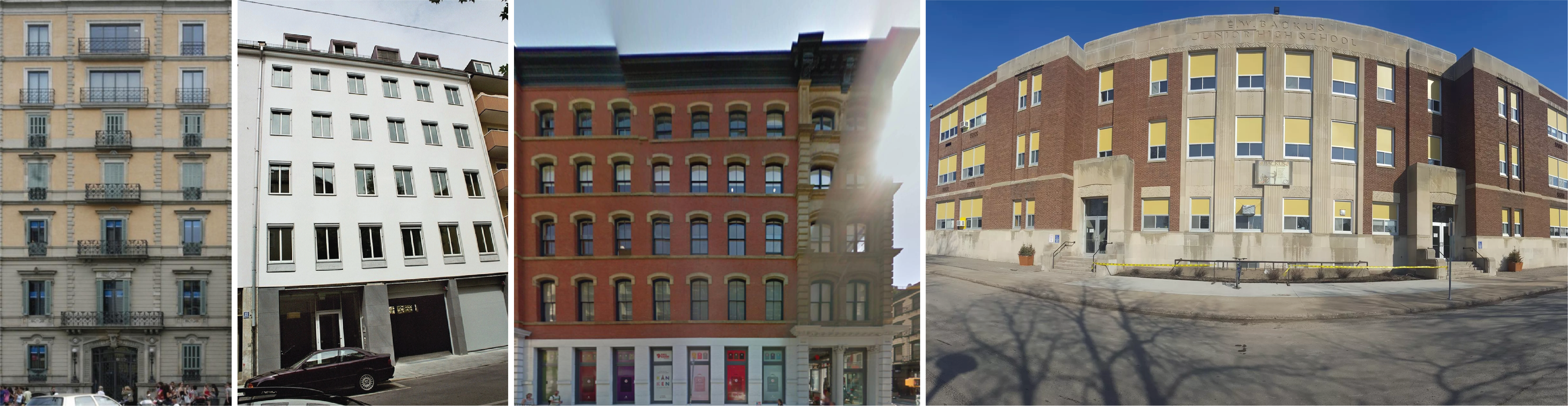}
    \caption{Different flavors of facade images considered in this paper; (left) a pre-rectified and cropped \facade from the CMP dataset; (center-left) a carefully acquired image from \eTRIMS, (center-right) a `wild' image automatically acquired from a street-view panorama image, including blooming and stitching artifacts; (right) photosphere panoramic image. }
    \label{fig:types-of-facade-image}
\end{figure*}

Typically, \facade images broadly come in three different flavors with increasing complexity (see \figref{fig:types-of-facade-image}). 
First, pre-rectified and cropped facade images, e.g.,~\cite{muller2007image}, which have been traditionally studied in \facade segmentation and parsing. 
Second, images that are not pre-rectified and  cropped but that are usually acquired with special care to have limited distortions. For example, images taken approximately from the front and containing single facades in the center covering most of the image. 
Third are input images acquired in the `wild,' particularly panoramic street-view images.
These are captured automatically, at scale, and then aligned and stitched together to form 360$^\circ$ photospheres.
We focus on the second and third types in this paper.

The advantage of `in-the-wild' images is that they are widely available and can be easily acquired automatically from a vehicle. However, the simplicity of data acquisition comes at the cost of increased challenges. 
For example, during capture no special attention is put on photographic details like viewpoint selection to avoid occlusions due to vegetation or passing cars, or unfavorable lighting conditions such as direct sun exposure.
Further, such panoramic street-view images contain many additional details besides buildings (e.g., foliage, scaffolding, vehicles, etc.). The focus of our work is to get the best possible facade segmentation and to extract architectural details from such raw images. 

In this paper, we present a deep learning based solution that is inspired by  recent success of semantic segmentation~\cite{badrinarayanan2015segnet,badrinarayanan2015segnet2,kendall2015bayesian}. We propose three new network architectures particularly focusing on \facade images. By explicitly accounting for the typical types of noise in  \facade datasets, enabling pixels to be assigned multiple labels, and exploiting the correlation among different \facade elements, we demonstrate significant improvement over existing state-of-the-art \facade segmentation methods. We evaluate our proposed method against a range of competing alternatives, and demonstrate significant improvements in terms of $F_1$ scores on multiple \facade benchmark datasets. 

\section{Related Work}
\label{sec:related-work}

\subsection{Traditional \facade parsing methods}

\Facade parsing or \facade segmentation was typically a mixture between traditional segmentation algorithms and finding ways to encode architectural priors.
Architectural priors can be encoded using grammars~\cite{muller2007image,riemenschneider2012irregular,teboul2013parsing}, symmetry~\cite{shen2011adaptive}, matrix rank~\cite{yang2012parsing}, MRFs~\cite{martinovic2012three,kozinski2015mrf}, CRFs~\cite{Mathias2016}, element templates~\cite{nan2015template,duygu:16:cgf}, rules~\cite{martinovic2012three}, hard constraints~\cite{cohen2014efficient}, and more general energy functions~\cite{dai2012learning,Mathias2016,jiang2016automatic}. The architectural information is typically encoded for architectural elements, e.g.,  rectangular regions in the segmentation with the same label. Some examples for architectural priors are element sizes and aspect ratios, allowable neighborhood relationships (e.g.,  chimney has to be on top of the roof), spacing between elements, constraints of alignment and size between elements (e.g.,  all windows in a floor need to be aligned and of the same height).
The low-level information can come from per-pixel classification algorithms, e.g.,  a boosted decision tree classifier~\cite{cohen2014efficient}, random forests~\cite{dai2012learning}, or mean-shift combined with recursive neural networks~\cite{martinovic2012three}. 
Multiple low level classifiers were evaluated in the ATLAS framework~\cite{Mathias2016}, 
but modern deep learning methods were not included. 
Alternately, it is possible to extract boxes of labeled regions using object 
detection algorithms~\cite{Mathias2016,Affara2016-ts,DeepFacade}.

One limitation of many traditional facade parsing methods is that they assume facade images that have been ortho-rectified and cropped (see \figref{fig:types-of-facade-image}-left). This allows to use much stronger architectural priors than facade parsing in the wild. For example, it is possible to make assumptions about shops being near the bottom of the image or windows being arranged in individual floors. Further, some data sets do not exhibit a strong variation in element arrangement and element size. For example the ECP dataset features many \facades{} of the same (Haussmanian) architectural style.

\subsection{Segmentation using CNNs}

Semantic segmentation is a classic topic in computer vision and has been heavily researched. In recent years, with the amazing success of deep learning, the state-of-the-art methods have produced large improvements. We refer the readers to a recent survey~\cite{semSeg_survey} for a summary of the current methods and details about typical accuracy measures. In the following, we particularly focus on learning methods specialized to \facade\ data.  

Schmitz and Mayer \cite{Schmitz2016-ou} used a \ac{CNN} to segment \facade images. They use `de-convolution' (also called transpose convolution) in order to up-sample the a CNN based on AlexNet, and they evaluated their results on the \eTRIMS dataset.

The `DeepFacade'\cite{DeepFacade} approach to facade segmentation used a fully convolutional net with a special loss function in order to segment \facade{} images. Their loss function penalized segmentation regions that were not horizontally and vertically symmetric.

However, our results indicate that a basic adaptation of general-purpose segmentation using \segnet~\cite{badrinarayanan2015segnet,badrinarayanan2015segnet2,kendall2015bayesian} is already better than the current state of the art deep learning methods. For example, Kelly et al.~\cite{Kelly:SIGA:2017} used SEGNET to determine the locations of windows, balconies, and doors in large scale procedural models of urban areas based on streetview imagery. We compare our results to retrained SEGNET and DeepFacade as the currently best available solutions for \facade image segmentation and report significant quantitative improvements.

\subsection{\Facade datasets}
The ECP dataset~\cite{Teboul2011-uy} contains 104 images of rectified and cropped facades with the label set \Target{window}, \Target{wall}, \Target{balcony}, \Target{door}, \Target{roof}, \Target{sky}, \Target{shop}.
The \eTRIMS dataset~\cite{korc-forstner-tr09-etrims} contains \numeTRIMSimages images with the label set \Target{building}, \Target{car}, \Target{door}, \Target{pavement}, \Target{road}, \Target{sky}, \Target{vegetation}, \Target{window}. 
These images are not rectified and not cropped, however, the images stem from a very careful viewpoint selection. 
All images have an almost frontal view of a single facade that fills most of the image.
The CMP dataset~\cite{Tylecek13} contains \numcmpimages annotated images with the label set \Target{facade}, \Target{molding}, \Target{cornice}, \Target{pillar}, \Target{window}, \Target{door}, \Target{sill}, \Target{blind}, \Target{balcony}, \Target{shop}, \Target{deco}, \Target{background}.

\section{Method}
\label{sec:overview}

%
%
Our goal is to classify \facade\ images into semantic pixel-level classes.
Although this is a special case of semantic segmentation, certain aspects of the target dataset make the problem unique.
First, the input images are often only partially rectified and suffer from various (unknown) camera and post-processing (e.g., stitching) artifacts. 
This makes it difficult to train for invariance under the difficult-to-model warping effects. 
Second, the desired features have vastly different proportions. 
For example, the windows and doors versus the ledges and window sills have very different aspect ratios. 
Finally, the typical \facade\ features share strong inter- and intra-label relations, which can easily get 
lost if their labeling tasks are considered in isolation.  

In our early experiments, we found that applying direct semantic segmentation pipelines result in fairly low $F_1$ scores especially on images such as \googlestreetview imagery. In our first attempt to build a \facade segmentation classifier, we trained on CMP imagery only; but the classifier appeared to generalize poorly to \googlestreetview imagery (based on initial qualitative evaluation). Therefore, we doubled the size of our training data with images we annotated from \googlestreetview and observed that \segnet{} is capable of giving competitive results when these images were included in our training and test sets, as indicated in row one of Table \ref{tab:quantitative-results-for-windows}. However, we still identified multiple modifications that can significantly improve upon the baseline \segnet{} approach for \facade segmentation~\cite{segnet2017}. All of our proposed modifications could, in principle, be applied to other architectures such as fully convolutional networks~\cite{Shelhamer2017-kq}, dilated nets~\cite{Yu2015-kw,Yu2017-yt}, or U-nets~\cite{Ronneberger2015-lx} as well, however we limit the scope of this work to \segnet.

\subsection{Data Augmentation}

Our goal is to segment images captured automatically, such as \googlestreetview imagery. This data poses several challenges because the images are the result of a number of processing and stitching steps that leave distortion artifacts in the images. Furthermore, the \ac{GPS} information associated with the images is imperfect, so \facade{}s are often not rectified or centered in the imagery; in fact there are often facades behind or around a central \facade{}. We address this as follows: 
\begin{itemize}
\item rectification using \cite{Affara2016-ts} to correct for errors in camera orientation; 
\item \emph{data augmentation} to force the classifier to be robust to errors in rectification by applying a random perspective warp to each image; and 
\item random sampling and manual labeling \googlestreetview imagery in the datasets used for training and testing.
\end{itemize}

\begin{figure}
    \centering
    \includegraphics[trim={0 4.2cm 0 2cm},clip,height=1in]{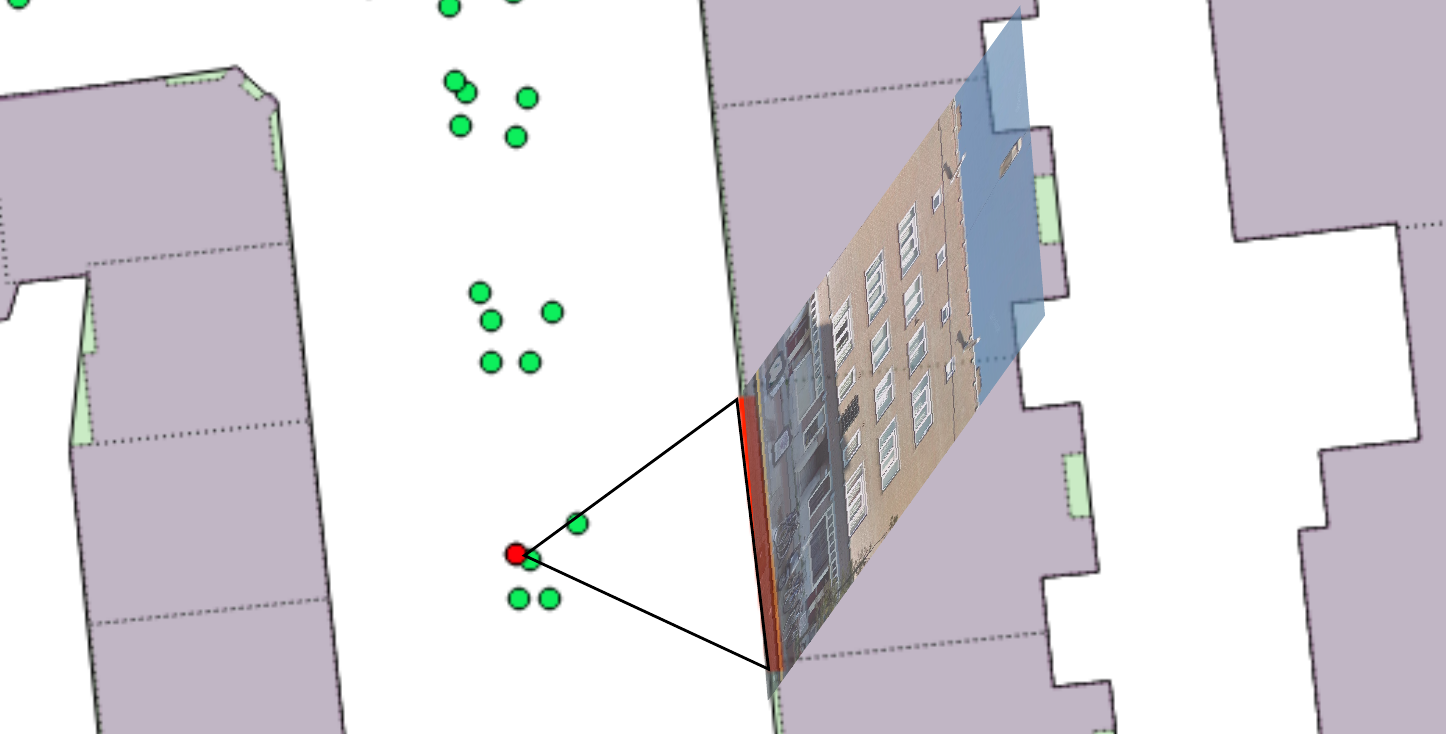}
    \includegraphics[height=1in]{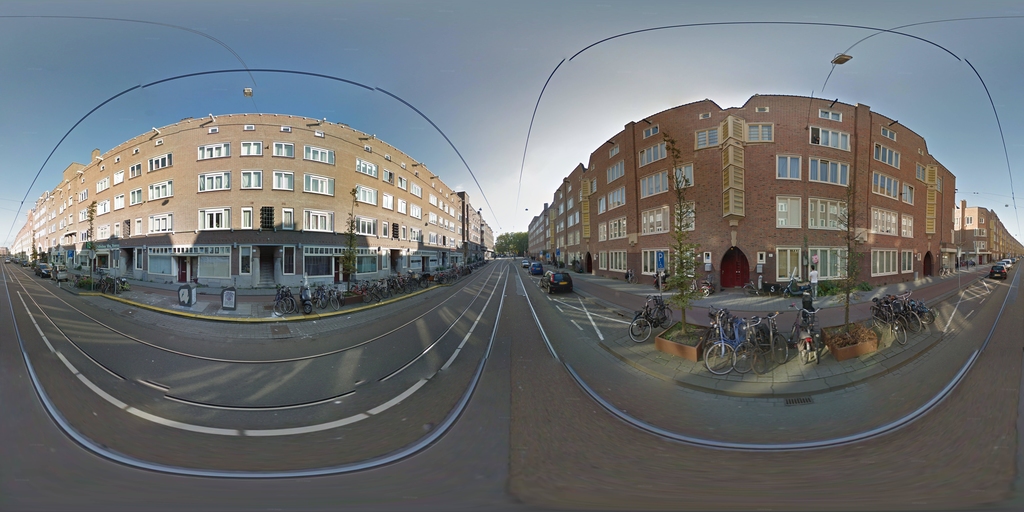}
    \caption{An example illustrating the way \ac{GIS} data is used to extract \googlestreetview imagery; shown (left) are building outlines from Amsterdam, Netherlands (violet, dotted lines), and their simplified and merged outlines (light green, solid lines). The location of \googlestreetview photospheres are indicated by green dots, and a selected wall and photo-sphere are indicated in red. The photosphere (right) is ray-cast onto a quad to form the \facade{} image.}
    \label{fig:gsv-facade}

\end{figure}

In order to collect \facade images we project \ac{GSV} \photosphere\ images onto planes derived 
from the linear segments of a \ac{GIS} polygonal building footprints
layer\footnote{The source code to download footprints  
will be made available online after the blind-review process is complete.} 
%
Building footprint datasets are quite common and can be obtained from public sites such as \openstreetmap; 
the training data we consider `from the wild' was extracted from \openstreetmap\ building footprints 
boundaries of 20 large metropolitan areas around the world 
(Amsterdam, Antwerp, Athens, Atlanta, Auckland, Austin, Berlin, Bern, Bordeaux, Bucharest,
Brisbane, Brussels, Cape Town, Chicago, Cleveland, Copenhagen, Dallas, Honolulu, and Hong Kong).
However, the building footprints are digitized at a variety of levels of precision; 
with some examples capturing sub-\facade{} details such as awnings or bay windows (see e.g., the upper left corner of \figref{fig:gsv-facade}).
We simplify and merge building footprints with a tolerance of \mergetolerance{}\meters\
in order capture the dominant plane of each \facade{} (or each group of collinear \facade{}s that form a wall of the building). 
We found \ac{GIS} height data for buildings to be inconsistent, and  treat each wall as though it were \nummeterstall{}\meters{} tall. 
The spatial resolution is also effected by the horizontal angles
between points on the \facade{} and rays towards the \photosphere{}, 
so we subdivide walls so that they are each approximately \nummeterslong{}\meters{} long.  
Each linear segment is extruded by \nummeterstall{}\meters{} upwards
in order to form a 3D quadrilateral in which \googlestreetview{} \photosphere{}s will be sampled. 
the quad is subdivided to form a grid of samples at a resolution of \metersperpixel{}\meters{} between samples. 
Finally, \facade images were generated by ray-casting from the \photosphere center of each  to each grid point, and using the colors where the ray intersects the \photosphere{}.

Based on the \photosphere{}'s odometry information, we find that our images are approximately rectified
(e.g., to within about 15\degrees{}).
However, the discrepancy is significant enough to obscure important horizontal 
and vertical alignments between features on the \facades{}. 
In order to account for errors in the orientation of each \photosphere{},  we extend each \facade{} by \nummetersexteded{}\meters{} before projecting. 
This also results in overlap between images on long walls, increasing the odds that each image contains a complete \facade{}. 
Then we use the single-image rectification approach of Affara \etal~\cite{Affara2016-ts} in order find a homography which increases the dominance of horizontal and vertical edges in each image.

For rectification, we use a classifier trained on \camvid{} data in order to identify pixels 
which are likely not part of a \facade{} (sky, pedestrians, vehicles, or vegetation) 
and we remove those edges from consideration for rectification. 
During training, each image is warped by a uniform random perspective transformation
which displaces the corners of the image by up to \warpamount{} percent of the image width. 

The segmentation approaches described in the following sections all share the same corpus of training data, which includes imagery from the CMP dataset and also labeled images captured from \googlestreetview{}. New data that was acquired was rectified, and then the boundaries of \facades{} were manually marked in each image. 
In a second phase, each individual \facade{} was extracted to form a single-\facade{} image, which was then completely labeled. 
During labeling, we encountered many out-of-model elements that were common in \facade{} images. 
We list \textit{window-AC units}, \textit{awnings}, \textit{fire escapes}, and \textit{bay-windows} as examples.
Our labeling process resulted in 22 common features, along with an `outlier' label and an `undeterminable' label for objects that we could not resolve. 
These include the 11 labels used for CMP and the 8 labels used in eTRIMS data which we use for comparisons.  
Of these labels, only the 11 CMP-labels were used for the model presented in this work; 
additional labels (\Target{sky}, \Target{roof}, and \Target{chimney}) were added 
for cross-validation models presented in Tables \ref{tab:ecp-five-fold-compatability} and \ref{tab:ecp-five-fold-precision-recall-f}.

\subsection{Baseline: \segnetfacade{}}

The \segnet \cite{segnet2017}  assigns labels to each
pixel using an auto-encoding approach; 
the VGG16 \cite{Simonyan2014-VGG} convnet architecture is used as \emph{encoding} layers
to form a deep representation of an input image and then a mirrored series
of \emph{decoding} layers that reverse the VGG max-pooling operations  
are used to reconstruct a dense output label image. 
\Segnet originally presented results on 
driving scenes from the \camvid \cite{BrostowFC:PRL2008:Camvid} 
data-set, but we re-targeted it to segment
\facade images using refinement 
learning.

The \segnet{} architecture was modified slightly to allow the input layer to 
accommodate $512\times512$ images, which was a compromise keeping the spatial resolution 
of the images large enough to resolve \facade{} elements, and keeping the images small enough to
fit within the available GPU RAM.   During training and inference, the 
input images were scaled so that their height was $512$ pixels, and after scaling they 
were partitioned into horizontally overlapping tiles that were each $512$ pixels wide.
Narrow images were padded with mirrored copies; if it was wider than $512$ 
pixels then it was tiled into the smallest number of $512$-wide tiles that overlap by at 
least $16$ pixels. After segmentation, the softmax scores for overlapping pixels were
averaged and then re-scaled before taking their argmax in order to determine the final label.
The loss function used by \segnet is a 
weighted cross entropy; the original weights were used to deal with class-imbalance issues so that
the weight of each label is inversely proportional to its frequency and the median frequency has a weight of one.
We computed new frequencies based our combined CMP+GSV dataset and used median-frequency class balancing when refining \segnet. 

In this paper, we refer to \segnet{} refined for \facade{} segmentation as \segnetfacade. 
Starting with \camvid model weights, we trained on 80\% of the labeled data (1200 tiles) using 
\ac{SGD} with weight decay and a low learning rate (1e-6) until the training loss 
plateaued (around 200 epochs). We continued to train the network until we reached
300 epochs.  

\subsection{Network Architecture 1: \segnetmultifacade}

\begin{figure}
    \centering
     \includegraphics[height=2in]{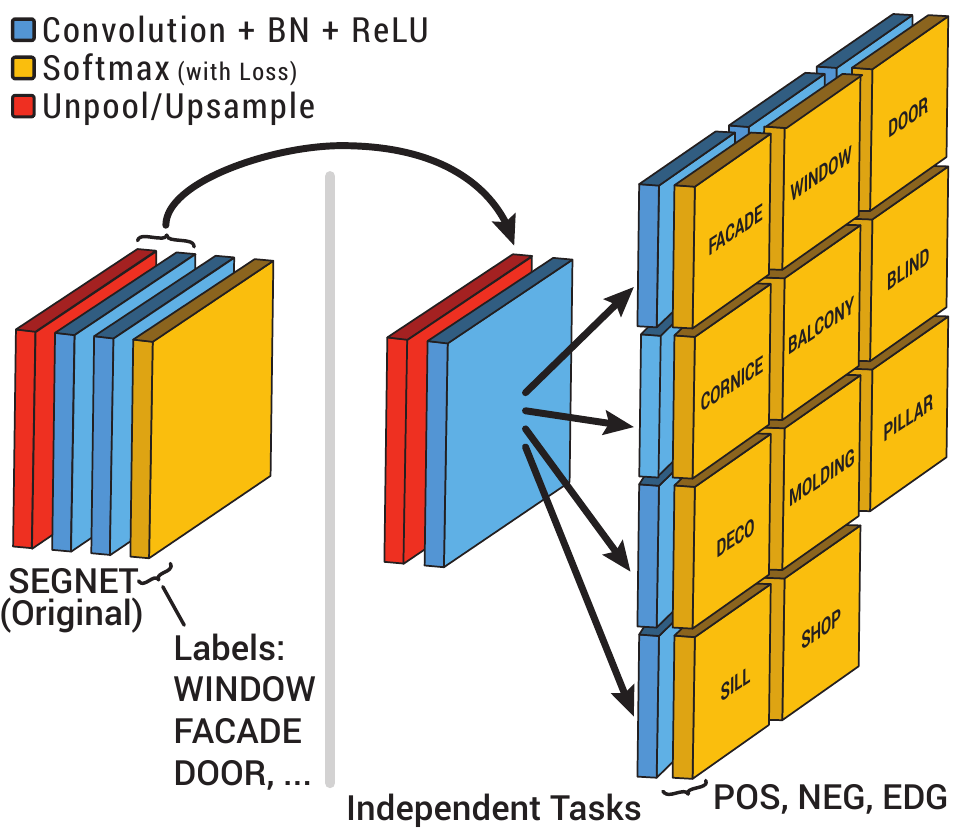}
    \caption{The new output layers for independent labeling. The last convolutional layer and softmax layer of \segnet{} are repeated \numlabels{} times; once per each type of object we aim to segment from \facade{} images. For each type of object we label pixels as \NEG{}, \POS{}, or \EDG{}, as wall as an additional \UNK{} label used to indicate lack of information during training (not used during inference).  }
    \label{fig:multi_label_architecture}
\end{figure}

\Facade segmentation is a multi-label problem where  each pixel can have multiple labels, i.e.,  the regions that are assigned to each label are not disjoint. 
For example, the regions of the image that are assigned the labels \windowlabel{} and \balconylabel{} often overlap.
In both the CMP dataset, as well as the data we created from \googlestreetview{} imagery, ground-truth annotations were provided as polygonal shapes that extend behind occluding features in the imagery so that we had access to multiple labels at each pixel for training. 
We conjectured that (i)~the task of labeling complete objects may be simpler than forcing the classifier to decide between two plausible labels when pixels are partially covered, and (ii)~the large receptive fields of each output of the net would allow it to recognize partially occluded objects if we did not force the output labels to be disjoint.  
Hence, we replaced the single \softmax{} operation of the \segnet{} classifier with \numlabels{} separate \softmax{} operations (we do assign a label to \Target{background} as it is considered a lack of any other label); treating each feature as a separate classification problem (See \figref{fig:multi_label_architecture}).

In principle, each feature could be treated as a separate \emph{binary} labeling problem, however, we use one additional label per feature. We observed that our baseline \segnetfacade segmentation had a tendency to produce smooth (or `blobby') outputs, and many facade features are thin, or are separated by thin regions in the image. It is important that the segmentation does not merge nearby objects, and it is also difficult for annotators to precisely mark the boundaries between objects. Furthermore, the \emph{boundaries}, or \emph{edges} of objects seem to take on different characteristics than their interiors; for example, the connected \windowlabel regions might be bounded by window frames.  We posit that object edges can be treated as a distinct class (based on their appearance) than the interiors of objects, and that treating edges as a separate target for each feature would drive our classifier to make more accurate predictions.

The \EDG{} label was assigned to pixels within 10cm of the edge of the target feature, before the images were scaled down to $512$ pixels in height. For certain features we decided that the edges should be handled differently; for the \facade{} itself we wanted to be able to determine the dividing line between adjacent \facades{} so we used one foot for the vertical edges. The tops and bottom of \facades{} were difficult to reliably label, so we did not mark horizontal edges for the \Target{facade} (a.k.a. \Target{wall}) element. 

The last decoding layer of \segnet{} was replaced by \numlabels{} different $3\times3$ convolutional layers corresponding to the  \numlabels{} CMP labels (excluding \Target{background}). 
Each convolutional layer had $4$ outputs, indicating whether the output pixel is \NEG{}, \UNK{},  \POS{} or an \EDG{} of the feature.
The class-imbalance issue among the four outputs is much more extreme than it was for training \segnetfacade{} because the \NEG{} label is far more frequent when a single element is considered in isolation. 
Instead of median-frequency balancing we opted to use assign a loss-weight of 1 to false \POS{} labels, 0.5 to false \NEG{} labels and a loss of 6 to false \EDG{} labels. 
These numbers reflect our best estimate of how important each type of error is. 

We initialized \segnetmultifacade classifier  using the 
baseline \segnetfacade weights, and trained using \ac{SGD} with weight decay
and a low learning rate (1e-6)  for 300 epochs. 

\subsection{Network Architecture 2: \separable}

We observe that \facade{} layouts are often approximately arranged on a flexible grid, with some exceptions. 
Furthermore, we operate on \facade{} images that are approximately rectified using an automatic method \cite{Affara2016-ts} 
so that most \facade{} elements occupy rectangular, nearly horizontally and vertically aligned regions.  
Therefore, we posit that we should be able to \emph{generate} \facade{} labels using convolutions by sequence of horizontal and vertical filters, which we expect to encourage synthesis of similarly aligned elements in the output.
Furthermore, since these filters require less parameters we can increase the lateral propagation of information between
labels in the output image by using much larger horizontal and vertical filters.
This could, for example, allow the decoding layer to  reconstruct details in the output labels that were occluded in the input.

\begin{figure}[t!]
    \centering
    \includegraphics[width=0.9\textwidth]{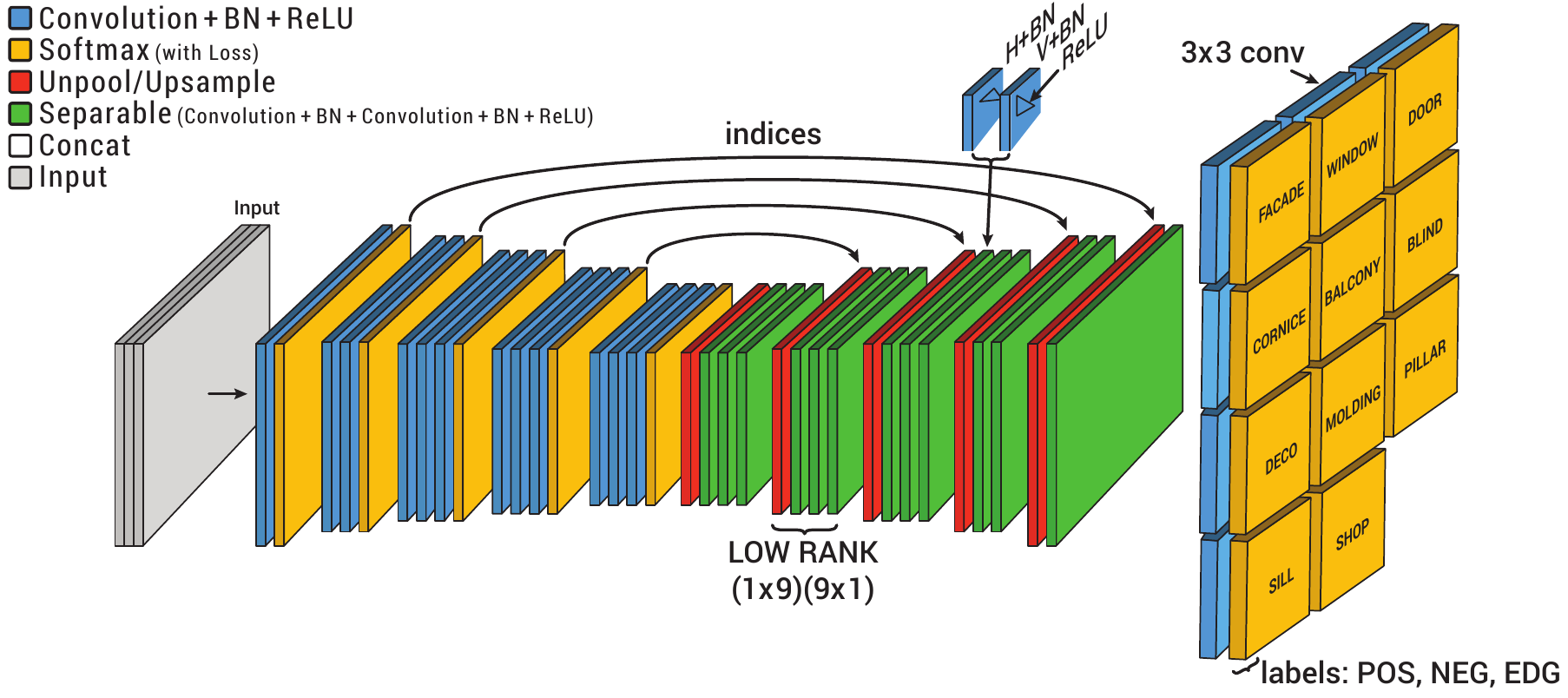}
    \caption{The \segnetseparable{} network architecture is formed by modifying the decoding layers of the \segnetmultifacade{} architecture and replacing each of the $3\time3$ convolutional layers (blue) with a pair of $1\times9$ horizontal and $9\times1$ vertical convolutions, each followed by batch normalization (pairs are indicated in green).}
    \label{fig:senget-separable-architecture}
\end{figure}

The \segnetmultifacade{} classifier used $3\times3$ convolutions during the decoding portion. We aim to replace them by a \textit{horizontal} convolution followed by a \textit{vertical} convolution, however there would be little effect if the filters were on $3$ pixels long. Instead we replaced each $3\time3$ filter with a $1\times9$ convolution, followed by a batch-normalization layer and $9\times1$ filter (See \figref{fig:senget-separable-architecture}).  This increased the number of trainable parameters in the decoding portion of the net, but we believe it also increased the spread of information as the output is hierarchically produced, so that long range relationships between output labels could be encouraged during the decoding process. 

The classifier was initialized using the \segnetmultifacade{} weights, with the new convolutional layers initialized using random values from a truncated standard normal distribution. We trained this network using \ac{SGD} with weight decay and a low learning rate (1e-6)  for 300 epochs.

\subsection{Network Architecture 3: \segnetcompatability{}}

\begin{figure}[t!]
    \centering
    \includegraphics[height=2.5in]{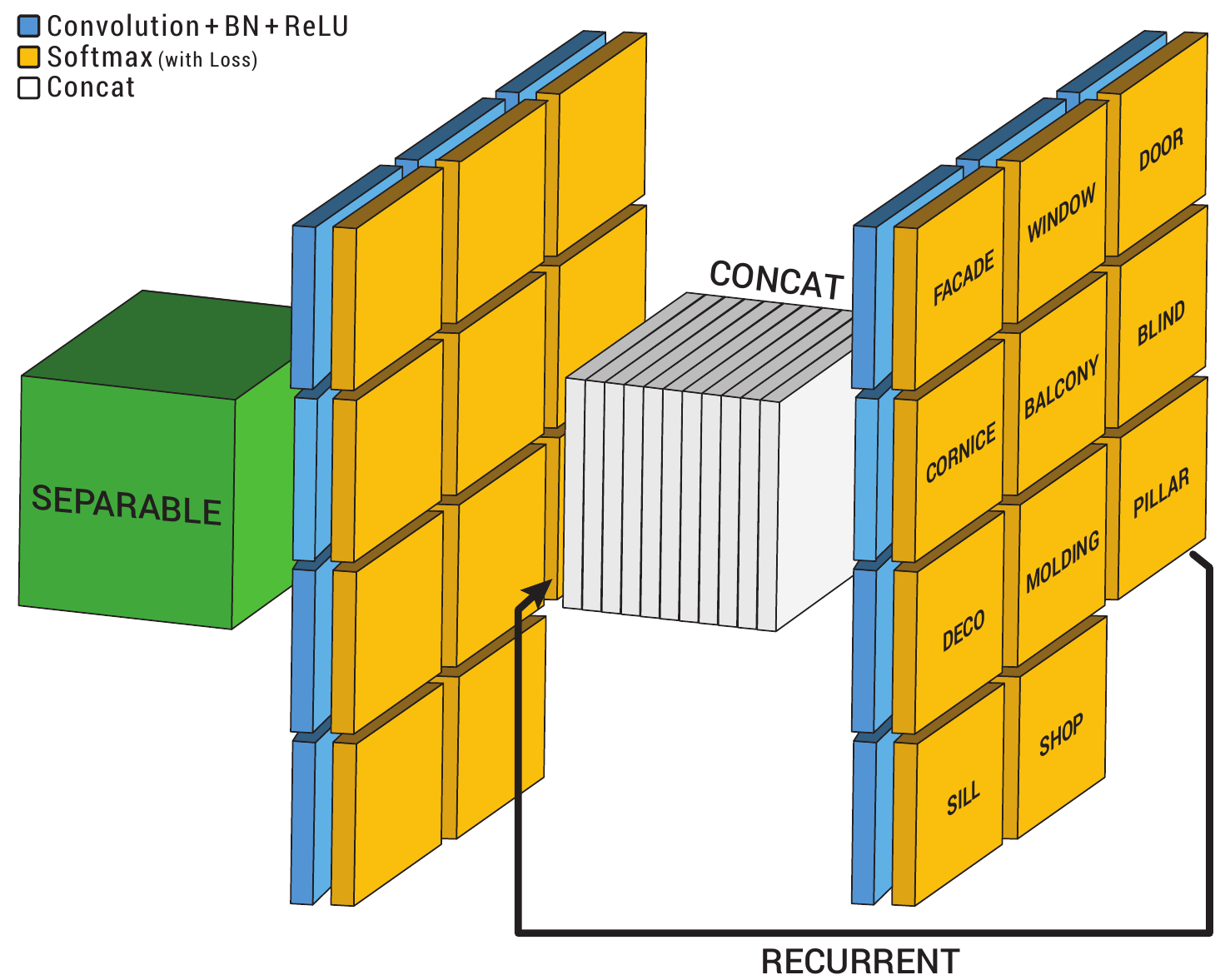}
    \caption{The changes made to \segnetseparable{} in order to become the \segnetcompatability{} architecture. The output of \segnetseparable{} is shown on the left, with softmax outputs (yellow) for each of the 11 labels. A \textit{recurrent block} of layers consisting of concatenation (white), $3\times3$ convolution (blue) and softmax nonlinearities (yellow) is appended to the network. Because the output and input are the same shape, this block can be repeated multiple times (e.g., twice in our experiments). }
    \label{fig:segnet-compatability-block}
\end{figure}

The \segnetseparable{} network for \facade{} segmentation decoupled the labels 
for the \numlabels different objects we aimed to identify in \facade{} images. 
However, for objects that are compatible with each other 
(e.g., \Target{shop} and \Target{window} or \Target{door}), 
whereas others are not. 
In addition, some objects (e.g.,  \Target{sill} or \Target{cornice}) are more likely to
occur in the vicinity of other classes of object such as \Target{window}.  
Much of this coupling is inherently captured by the large receptive fields
and information sharing that happens as \segnet{} encodes and then 
decodes an image, however we suspect that certain errors that are indicated 
by incompatible labels being used together could best
be identified from the outputs of a segmentation approach. 
In order to address this possibility we created a recurrent block (see \figref{fig:segnet-compatability-block})
of output layers that follow the output of \segnetseparable{}. 
Each block starts with a concatenation of the softmax outputs of \segnetseparable{}, followed by \numlabels different $3\times3$ convolutional layers corresponding to the  \numlabels{} output labels, each taking the entire concatenated layer as input. Each convolutional layer is followed by another softmax operation (which adds a non-linearity to the process). The entire block produces output that is the same shape and semantics as its input, so one could repeat the block any number of times in a recurrent fashion.

Specifically, we added two blocks (concatenation, \numlabels convolutions, and softmax) to the trained \segnetseparable{} network, essentially unrolling the loop in the network architecture twice for training. Since the recurrent convolutional layers in \figref{fig:segnet-compatability-block} occur twice in the unrolled network, we ensured that the weights were shared between convolutional layers each time they were repeated. During training, we added together the weighted cross-entropy losses associated with each softmax unit, so if the compatibility block is repeated twice there are a total of $\numlabels{}\times3=33$ loss terms associated with the output. In order to prevent random initialization of weights in the new convolutional layers from adding noise to the earlier layers of the net, we first froze all of the weights in \segnetseparable{} and trained the network for 100K iterations (it converged much more quickly) with a learning rate of $1e-4$. Then we restored the learning rate for the initial \segnetseparable{} layers of the net and resumed training for another 100K iterations in order to produce a final \segnetcompatability{} network. Training this network proved to be difficult; we had to increase the learning rate by a factor of 100 for the recurrent layers and set a negative slope to all (leaky)~ReLU activation to prevent the network training from stalling. 

\section{Results}
\label{sec:results}

We evaluate our approaches using a holdout set of \numholdouts{} images, with \numcmpholdouts holdout images from the CMP \cite{Tylecek2012-CMP,Tylecek2013-CMP} dataset of around 600 rectified facade images, the remaining \numourholdouts{} where randomly selected \googlestreetview images that we had annotated for this project.
In order to quantify and compare our results we use the accuracy ($\pxacc{}$) as well as the precision ($\pxprecision{}$), recall $(\pxrecall{})$, and $F_1$-measure $(\pxfscore{})$. Although accuracy is often reported, we found that $F_1$-score seems to correlate with our visual impression of the quality of the result.

\begin{table*}[t!]
    \caption{Comparison on \numeTRIMSimages\ images from \eTRIMS for \windowlabel{} based 
    on our \separable{} model.}
    \label{tab:eTRIMS-quantitative-results-for-windows}
    \centering
         \begin{tabular}{lccccccc}\hline
        Approach  & \pxacc & \pxprecision & \pxrecall & \pxfscore\\
        \hline 
        Yang and F\"orstner 2011~\cite{Yang2011-zc} &   0.75 &   0.75   & 0.60  & 0.67  \\
        ATLAS~\cite{Mathias2016-xc} &   0.73 &   -.--   & -.-- & -.-- \\
        Cohen \etal 2014~\cite{Cohen2014-nl} &   0.71 &   -.--   & -.--  & -.-- \\
        Schmits and Mayer 2016~\cite{Schmitz2016-ou}  &   0.86 &   0.67   & 0.71  & 0.69 \\
        DeepFacade~\cite{DeepFacade}   &   0.91 &   -.--   & -.-- & -.-- \\
        \hline                  
        Ours         &   0.971  &   0.89    & 0.64  & 0.74  \\ \hline
        \end{tabular}
        \vspace*{-.2in}
\end{table*}



We introduced the class \EDG{} into our labeling scheme during training,
but during inference and testing we exclude the \EDG{} label 
by normalizing the \POS{} and \NEG{} probability outputs of our classifier to sum to one.
Pixels marked as \UNK{} or \EDG{} in our ground-truth data are ignored; 
we consider ignoring \EDG{} to be a reasonable decision 
as annotators are often uncertain about the precise locations 
of object-boundaries \cite{Everingham2010-yo},
however, ignoring edges seems to have little effect 
on the numbers. 
When we compare against other methods in Table \ref{tab:ecp-five-fold-compatability} 
we do \textit{not} ignore \EDG{} labels 
in order to ensure that comparisons are fair to prior art.


For object based scores, we find the bounding boxes of each connected component in our argmax outputs, and in the ground truth. The object score, especially precision, can be heavily influenced by small spurious components for regions of pixels near the classifier's decision boundary. This could be addressed using \ac{CRF} optimization but we instead use a small $3\times3$ morphological opening operation prior to connected component labeling.
We consider two objects to be a potential match if their \ac{IoU} is more than 0.5, and we find a maximum weighted bipartite matching between objects detected as positive by our system, and objects from the ground truth. We consider matching objects to be true-positives ($\obTP$) and the umatched objects are false alarms ($\obFP$) and misses ($\obFN$). We also report object-based recall ($\obrecall{}$), precision ($\obprecision{}$) and $F_1$-scores ($\obfscore{}$) . 

\begin{table*}[b!]
    \caption{Quantitative comparison on ECP data using our \separable{} model. We used five-fold cross validation on ECP and we show the mean and variance across folds, the top scores are indicated in bold. Following \cite{DeepFacade}, we show results compared to several approaches; (1) is Yang and F\"orstner \cite{Yang2011-zc}, (2.1, 2.2) are two variants of Mathias \etal \cite{Mathias2016}, (3.1, 3.2, 3.3) are three variants of Cohen \etal \cite{Cohen2014-nl}, and (4) is the best of three variants presented by Liu \etal \cite{DeepFacade}.} 
    \label{tab:ecp-five-fold-compatability}
    \centering
    \begin{tabular}{lcccccccc}\hline
        Class                   & (1)  & (2.1)   & (2.2)  & (3.1)  & (3.2)   & (3.3)   & (4) & \textbf{Ours}\\ \hline

        \Target{window }        & 62    & 76    & 78    & 68    & 87    & 85    & 93.04            & \textbf{$\mathbf{95.6 \pm 0.23}$}\\
        \Target{facade(wall) }  & 82    & 90    & 89    & 92    & 88    & 90    & \textbf{96.14 }           & $91.70 \pm 0.39$\\
        \Target{balcony}        & 58    & 81    & 87    & 82    & 92    & 91    & 95.07            & \textbf{$\mathbf{96.0 \pm 0.25}$}\\
        \Target{door   }        & 47    & 58    & 71    & 42    & 82    & 79    & 90.95            & \textbf{$\mathbf{98.8 \pm 0.09}$}\\
        \Target{roof   }        & 66    & 87    & 79    & 85    & 92    & 91    & 94.02            & \textbf{$\mathbf{97.7 \pm 0.10}$}\\
        \Target{chimney}        & -     & -     & -     & 54    & 90    & 85    & 91.30            & \textbf{$\mathbf{98.9 \pm 0.10}$}\\
        \Target{sky    }        & 95    & 94    & 96    & 93    & 93    & 94    & 97.72            & \textbf{$\mathbf{98.4 \pm 0.12}$}\\
        \Target{shop   }        & 88    & 97    & 95    & 94    & 96    & 94    & 95.68            & \textbf{$\mathbf{96.9 \pm 0.26}$}\\
        \hline
        total acc.              & 74.71 & 88.07 & 88.02 & 86.71 & 89.90 & 90.34 & 95.40            & \textbf{96.74}\\
        \hline
    \end{tabular}
\end{table*}



\begin{figure*}[t!]
    \centering
    \includegraphics[width=\textwidth]{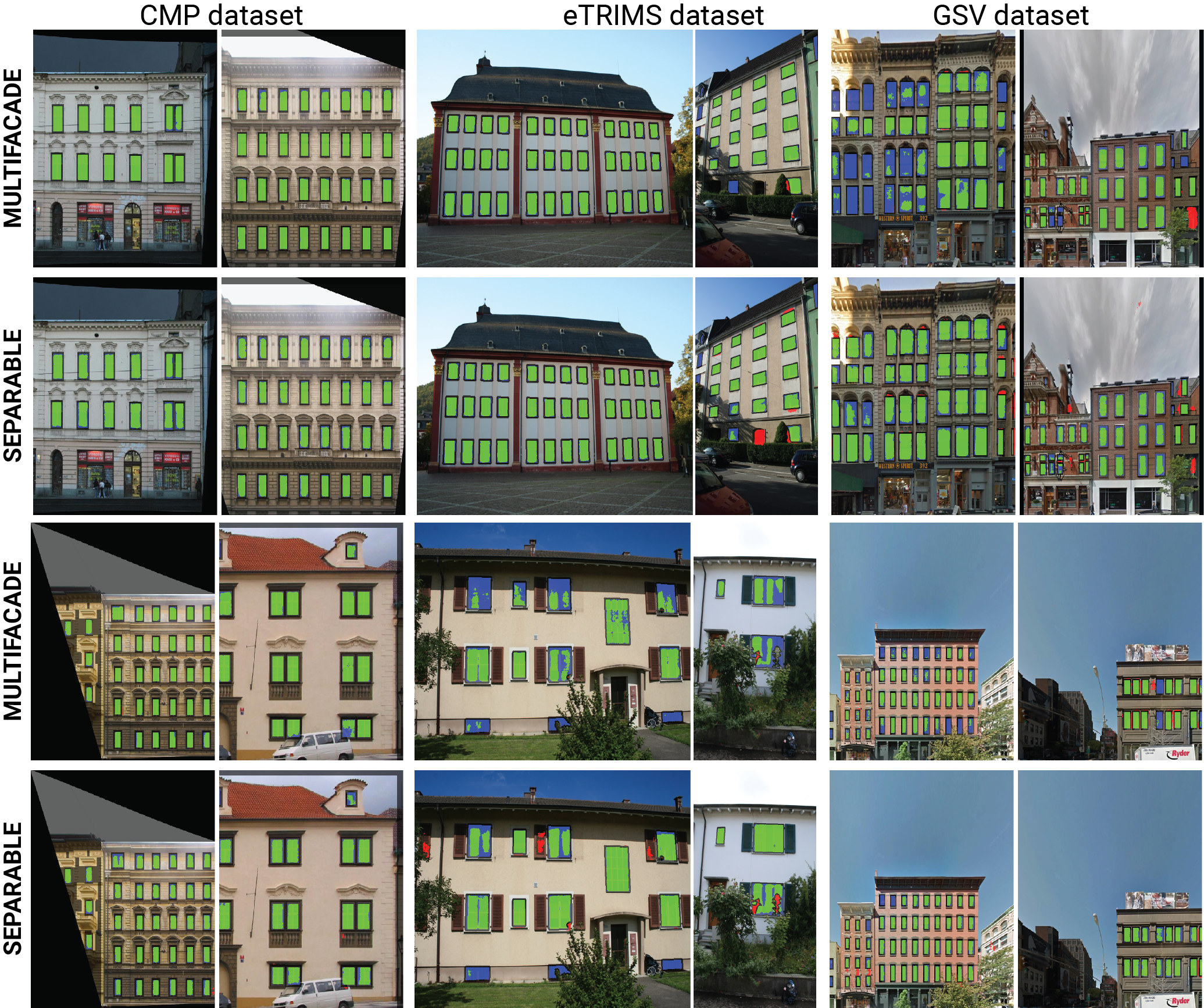} 
    \caption{Examples of estimated probabilities assigned to the \Target{window} class for a variety of images in our evaluation sets. }
    \label{fig:window-probabilities}
\end{figure*}

\begin{table}
	\caption{Accuracy, Precision, Recall, and $F_1$ for \separable{} on ECP data based on 5-fold cross validation. We suggest that accuracy, which includes the true-negatives, is a poor way to evaluate \facade{} segmentation as it rewards rare objects, for example compare it to the $F_1$ scores for \Target{door} and \Target{chimney}. }
	\label{tab:ecp-five-fold-precision-recall-f}
	\centering
	\begin{tabular}{l c@{\hskip 3ex}c@{\hskip 3ex}c@{\hskip 3ex}c}\hline
		label                  &       Accuracy $(A)$             &     Precision  $(P)$             &    Recall   $(R)$            &        $F_1$         \\ 
		\hline  
		\Target{window }       & $95.6 \pm 0.23$     & $81.5 \pm 2.03$     & $79.1 \pm 1.58$    &  $80.4 \pm 0.90$ \\  
		\Target{facade(wall)}  & $91.7 \pm 0.39$     & $88.6 \pm 1.17$     & $93.4 \pm 0.70$    &  $90.9 \pm 0.40$ \\  
		\Target{balcony}       & $96.0 \pm 0.25$     & $86.8 \pm 1.06$     & $83.2 \pm 3.11$    &  $84.9 \pm 1.61$ \\  
		\Target{door   }       & $98.8 \pm 0.09$     & $49.8 \pm 4.91$     & $53.5 \pm 4.76$    &  $49.9 \pm 2.68$ \\  
		\Target{roof   }       & $97.7 \pm 0.10$     & $84.6 \pm 0.75$     & $78.2 \pm 3.10$    &  $81.1 \pm 1.45$ \\  
		\Target{chimney}       & $98.9 \pm 0.10$     & $67.6 \pm 3.41$     & $61.4 \pm 4.43$    &  $64.0 \pm 2.87$ \\  
		\Target{sky    }       & $98.4 \pm 0.12$     & $83.6 \pm 2.61$     & $94.7 \pm 0.84$    &  $88.8 \pm 1.28$ \\  
		\Target{shop   }       & $96.9 \pm 0.26$     & $94.1 \pm 2.55$     & $83.2 \pm 3.23$    &  $88.1 \pm 0.97$ \\  
		\hline
	\end{tabular}
\end{table}

\begin{table*}
    \caption{Quantitative results for \windowlabel{}, based on \numcmpholdouts CMP holdout images (of \numcmpimages total).}
    \label{tab:cmp-only-quantitative-results-for-windows}
    \centering
         \begin{tabular}{lccccccc} \hline
        Approach             & \pxacc   & \pxprecision & \pxrecall & \pxfscore & \obprecision & \obrecall & \obfscore  \\
        \hline
        \segnet{}            &     0.93 &   0.74      & 0.70      & 0.72 & 0.72 & 0.67 & 0.69  \\
        \segnetmultifacade{} &     0.94 &   0.98      & 0.57      & 0.72 & 0.82 & 0.66 & 0.73  \\
        \separable{}         &     0.96 &   0.96      & 0.71      & 0.81 & 0.81 & 0.71 & 0.76  \\
       \segnetcompatability{}&     0.95  &   0.80      & 0.86      & 0.83 & 0.81 & 0.74 & 0.77  \\
        \hline
        \end{tabular}
\end{table*}

\begin{table*}
    \caption{Quantitative results for the \Target{window} class on \googlestreetview data with focus on the $F_1$-scores as a predictor of performance; each modification has led to a substantial increase in $F_1$.}
    \label{tab:quantitative-results-for-windows}
    \centering
     \begin{tabular}{lccccccc}\hline
    Variant
                          & \pxacc & \pxprecision & \pxrecall & \pxfscore & \obprecision & \obrecall & \obfscore  \\ \hline
    \segnet{}             & 0.93  & 0.55         & 0.62      & 0.58      & 0.76         & 0.57      & 0.65\\
    \segnetmultifacade{}  & 0.96  & 0.92         & 0.56      & 0.70      & 0.88         & 0.61      & 0.72 \\
    \separable{}          & 0.97  & 0.81         & 0.70      & 0.75      & 0.86         & 0.70      & 0.77 \\
    \segnetcompatability{}  & 0.95 & 0.85         & 0.79      & 0.82      & 0.81         & 0.72      & 0.76\\    \hline
    \end{tabular}
\end{table*}

\begin{table*}
    \caption{Quantitative results for all labels on CMP + GSV (combined) data using \separable.}
    \label{tab:cmp-and-gsv-all-labels-with-separable}
    \centering
    \begin{tabular}{lccccccc}\hline
     Target  & Acc & $P$ & $R$ & $F_1$ & $P_{ob}$ & $R_{ob}$ & $F_{1ob}$  \\
    \hline
    \texttt{ balcony } & 0.97 & 0.79 & 0.51 & 0.62 & 0.34 & 0.45 & 0.39\\
    \texttt{ blind } & 0.98 & 0.63 & 0.22 & 0.33 & 0.35 & 0.21 & 0.26\\
    \texttt{ cornice } & 0.98 & 0.73 & 0.55 & 0.63 & 0.43 & 0.33 & 0.38\\
    \texttt{ deco } & 0.98 & 0.43 & 0.15 & 0.23 & 0.30 & 0.08 & 0.12\\
    \texttt{ door } & 0.99 & 0.39 & 0.49 & 0.43 & 0.15 & 0.41 & 0.22\\
    \texttt{ molding } & 0.94 & 0.90 & 0.53 & 0.67 & 0.21 & 0.42 & 0.28\\
    \texttt{ pillar } & 0.99 & 0.75 & 0.00 & 0.01 & 0.60 & 0.00 & 0.01\\
    \texttt{ shop } & 0.97 & 0.46 & 0.69 & 0.55 & 0.11 & 0.15 & 0.13\\
    \texttt{ sill } & 0.98 & 0.72 & 0.21 & 0.32 & 0.22 & 0.11 & 0.15\\
    \texttt{ window } & 0.96 & 0.93 & 0.74 & 0.82 & 0.79 & 0.75 & 0.77\\
    \hline
    \end{tabular}
\end{table*}



\begin{figure*}[t!]
    \centering
     \includegraphics[width=0.48\textwidth]{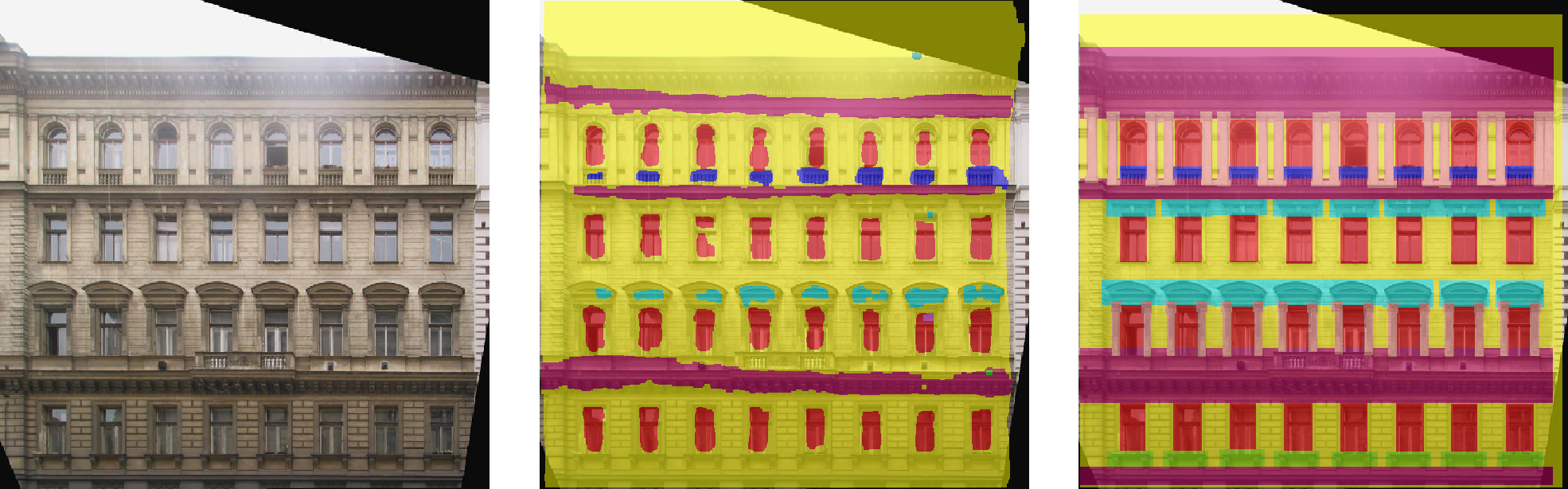}  
    \hfill 
    \includegraphics[width=0.48\textwidth]{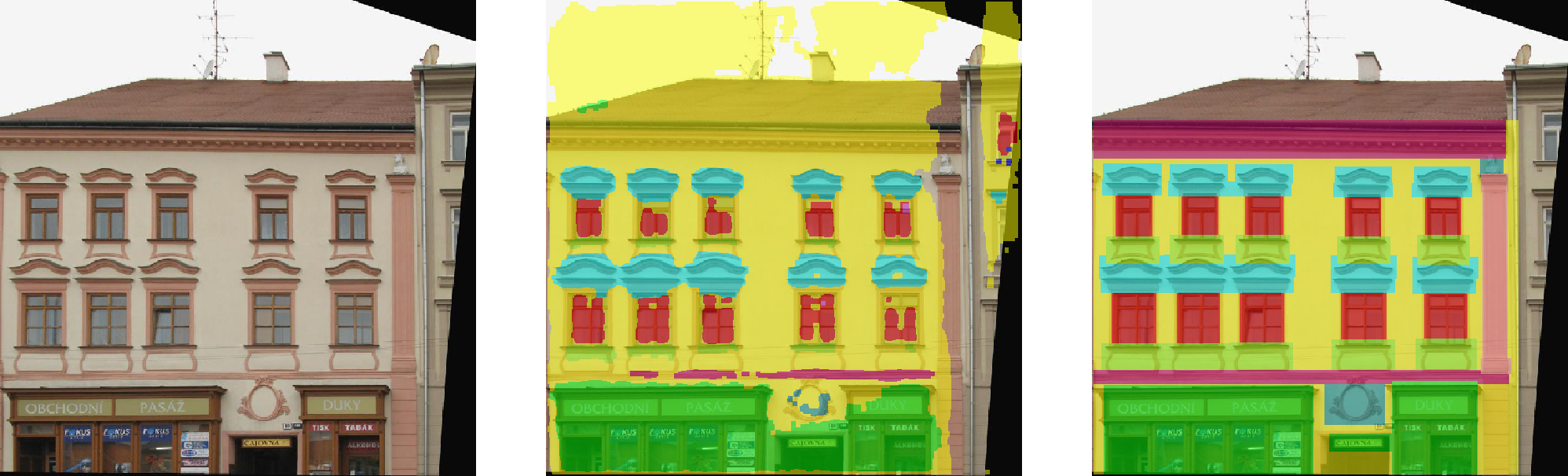}
    \caption{Example estimated probabilities for different \facade elements using \separable network. }
    \label{fig:facadeElementTypes-probabilities}
\end{figure*}

In our evaluations, we expect annotators to be precise to within a 5-pixel (10cm) boundary around the edges of the labeled regions.
Labels within this boundary region were excluded from our evaluation as they are unreliable annotations; 
this is the same approach taken  for example in VOC challenge data \cite{Everingham2010-yo}.
Table \ref{tab:cmp-only-quantitative-results-for-windows} shows our results for \windowlabel{} on the CMP data, and Table \ref{tab:quantitative-results-for-windows} shows our results for \windowlabel{} on our street view dataset. We can observe that our new additions to the \segnet architecture provide significant improvements over a baseline method \segnetfacade{}.
We can observe that both \segnetmultifacade{}, as well as \separable{} lead to better results; \segnetcompatability{} is sometimes best. In Fig.~\ref{fig:window-probabilities} we show a visual comparison between different variants and in Fig.~\ref{fig:facadeElementTypes-probabilities} we show visual results for different labels.

\mypara{Comparison on \eTRIMS{}}
In addition, we compare against a number of other \facade{} segmentation approaches using the \eTRIMS{} data set. Table \ref{tab:eTRIMS-quantitative-results-for-windows} shows results on \eTRIMS{} data, however their labels are not in perfect semantic agreement with the labels used to train our network; in particular partially occluded windows are considered negative examples, and our \shoplabel{} label is considered to be \windowlabel{} in \eTRIMS. Nevertheless our accuracy is higher than other reported accuracies, as is our $F_1$-score on this dataset. 
Unlike approaches that used cross-validation on \eTRIMS to generate their results, our results are based on a network that was trained of none of the \eTRIMS{} data which makes our result significant; although our evaluation is limited to only windows. 
The main point of this evaluation is to show that our results are still very good and that even our baseline \segnetfacade{} algorithm is already better than other state of the art approaches for facade segmentation. In particular, baseline \segnetfacade{} outperforms other deep learning based approaches.

\mypara{Comparison against ECP}
A number of authors report accuracy on the ECP dataset, so we refine our network to do 5-fold cross validiation on ECP imagery.  We added new outputs (increasing the number from 11 to 15) for ECP elements not in our original training set. ECP data did not include labels for overlapping objects, so in our multi-label representation of ground-truth we marked occluded regions as \UNK\ during training. For evaluation, we composite each the of 8 ECP labels to create an image that matches the ECP ground-truth format. Quantitative results are presented in Table \ref{tab:ecp-five-fold-compatability} and compared with a variety of approaches; for most features (with the exception of \Target{facade}) our \segnetseparable{} approach outperforms other methods according to accuracy. We also report precision, recall, and $F_1$-score per-label in Table \ref{tab:ecp-five-fold-precision-recall-f}. 

\mypara{Comparison between approaches}
Our aim is to find an approach with a high $F_1$-score in relatively unconstrained images (e.g. \googlestreetview imagery), and in particular with high per-object recall $\obrecall{}$ and $\obfscore{}$ because we are inspired by inverse procedural modeling. In Table \ref{tab:cmp-only-quantitative-results-for-windows} we demonstrate that each approach significantly increased the $F_1$ and $\obfscore{}$ on our most important class of object (\Target{window}), and similarly when we limit the results to only \googlestreetview imagery in Table \ref{tab:quantitative-results-for-windows} with the exception that the \segnetcompatability{} variant hurt our object-based scores.   In order to understand how these results break down between labels, we report per-element pixel and object-based metrics for all hold-outs in Table \ref{tab:cmp-and-gsv-all-labels-with-separable} using the \separable{} approach.  On a per-pixel basis we achieve satisfying results, however our simplistic approach identifying objects based on those pixels only leads to satisfying results on windows.

\section{Conclusion}

We presented a deep-learning based facade segmentation approach that works for a variety of data sets from pre-rectified and cropped facade images to automatically captured panoramic street view images. Starting from the established \segnet architecture, our main ideas are to add separate edge labels, overlapping labels, separable filters, and an iterative optimization for label smoothing. 
We also provide a larger dataset consisting of labeled street level photospheres. 
Our results demonstrate a significant improvement over the state of the art.
In future work, we would like to extend the work to improve object detection especially for thin objects, 
to investigate an end-to-end network trained for rectification and segmentation using transformer networks as well as employing separable filters for region completion (e.g., due to occlusion from trees).  

\clearpage

\bibliographystyle{splncs}
\bibliography{references}

\end{document}